\documentclass[runningheads]{llncs}
\usepackage[T1]{fontenc}

\usepackage{graphicx}

\usepackage{xcolor}
\usepackage{csvsimple}
\usepackage{amssymb,amsmath}
\usepackage{algorithm}
\usepackage{bm}
\usepackage[noend]{algpseudocode}
\usepackage[export]{adjustbox}
\usepackage{enumitem}
\usepackage{hyperref}
\usepackage{multirow}
\usepackage{multicol}
\usepackage{subcaption}
\usepackage{tabularx}
\usepackage{fixltx2e}
\usepackage{diagbox}

\usepackage{booktabs}
\usepackage{mathtools}
\DeclarePairedDelimiter\ceil{\lceil}{\rceil}

\usepackage{fixltx2e}
\usepackage{diagbox}

\usepackage{makecell}

\newcommand{\Mod}[1]{\ (\mathrm{mod}\ #1)}

\begin{document}
\title{Chains of Autoreplicative Random Forests for missing value imputation in high-dimensional datasets}
\titlerunning{Chains of Autoreplicative Random Forests}
%
\author{Ekaterina Antonenko\inst{1,2} \and Jesse Read\inst{1}}
\authorrunning{E. Antonenko and J. Read.}
%
\institute{LIX, \'Ecole Polytechnique, Institut Polytechnique de Paris, France \and
Digitalent lab (Moteur Intelligence Artificielle), Paris, France
\email{\{ekaterina.antonenko,jesse.read\}@polytechnique.edu}}
\maketitle              
\begin{abstract}

Missing values are a common problem in data science and machine learning. Removing instances with missing values can adversely affect the quality of further data analysis. This is exacerbated when there are relatively many more features than instances, and thus the proportion of affected instances is high. Such a scenario is common in many important domains, for example, single nucleotide polymorphism (SNP) datasets provide a large number of features over a genome for a relatively small number of individuals. To preserve as much information as possible prior to modeling, a rigorous imputation scheme is acutely needed. 
While Denoising Autoencoders is a state-of-the-art method for imputation in high-dimensional data, they still require enough complete cases to be trained on which is often not available in real-world problems.
In this paper, we consider missing value imputation as a multi-label classification problem and propose Chains of Autoreplicative Random Forests. Using multi-label Random Forests instead of neural networks works well for low-sampled data as there are fewer parameters to optimize. Experiments on several  SNP datasets show that our algorithm effectively imputes missing values based only on information from the dataset and exhibits better performance than standard algorithms that do not require any additional information. In this paper, the algorithm is implemented specifically for SNP data, but it can easily be adapted for other cases of missing value imputation.

\end{abstract}
%
%
%

\section{Introduction}

Missing values are a common problem and an important issue in the domain of data science and machine learning. Most off-the-shelf statistical and machine learning methods cannot handle missing values, and such values must be imputed, or the whole instance or row removed, before the actual data analysis. When many values are missing, considering only instances with complete information can lead to a loss of necessary information and can yield a very poor or even empty dataset.

A special challenge is missing values occurring in several or even most of the samples and/or features of the training set, and when there are sufficiently more features than samples ($p \gg N$), which means that removing samples amplifies the imbalance.
Single Nucleotide Polymorphism (SNP) is an example of high-dimensional and low-sampled categorical data where missing values are very common and can affect a large number of the features. Other examples of data with such characteristics include gene expression arrays, classification problems in astronomy, tool development for
finance data, and weather prediction~\cite{p>>N}.


Multiple Imputation with Chained Equations (MICE)~\cite{MICE} remains a state-of-the-art approach in the imputation domain, very powerful and flexible, but requires proper parameter tuning and a deep understanding of the data. We have not seen evidence of MICE usage for high-dimensional data. We propose possible MICE parameters to make computation time feasible but do not obtain promising results. Denoising Autoencoders have proved to work well for the missing value imputation~\cite{SCDA}, but they require enough complete cases for the training phase, which is not always the case in real-world data and, in particular, high-dimensional data.

In this paper, we consider missing value imputation of categorical features as a multi-label classification problem, and thus exploiting multi-label models such as Random Forests~\cite{RF} becomes possible. We present
an algorithm that
efficiently imputes missing values when data has complete cases to be trained on and can be adapted for high-dimensional data when having fully complete cases is very unlikely.
We explore the idea of a multi-label prediction cascade, using a methodology similar to that of, e.g., Classifier Chains~\cite{CCReview} in the multi-label classification literature, where already processed targets are stacked as new features for further targets. We use Chains of Autoreplicative Random Forests (ChARF) such that each Random Forest processes one window of consequent features and incorporates information from already imputed 
previous windows. We treat windows as data that can be given to an Autoencoder, however, noticing that we do not explicitly need a hidden-layer representation, we use multi-label Random Forests instead of neural network architecture. 
To the best of our knowledge, using simpler multi-label models as Autoencoders without explicit encoding has not been widely studied in the literature. We argue that such an approach can be useful for similar purposes, at least for imputation (similarly to Denoising Autoencoders), and has its advantages, such as an ability to efficiently process data of small sample size.

We study imputation for SNP data as it exhibits all the aspects we are interested in tackling: high dimensional data (such that $p \gg N$), the possibility of a significant proportion of missing values, and no reference panel but at least some local correlations in the feature space. 
At the same time, our approach can be adapted to any data exhibiting these characteristics. 
Although SNP data is an instance of categorical data, our model can easily be modified to work with continuous data by using regressor models, which have already been investigated in the context of `regressor chains', e.g.,~\cite{KatiaMultiModal}.
For high-dimensional and low-sampled datasets, the ChARF method shows to be very competitive w.r.t. both imputation quality and time complexity. At the same time, even for low-dimensional data we can adapt this approach with personalized splitting for blocks depending on missing patterns.

The rest of the paper is organized as follows. After summarizing background and related work in Section~\ref{sec:background}, we present our method in Section~\ref{sec:method}. The results and their discussion as well as complexity analysis are in Section~\ref{sec:results}. In Section~\ref{sec:conclusions}, we draw conclusions and describe future work.


\section{Related work}
\label{sec:background}


Traditionally, missing data is categorized into three types:
Missing Completely At Random (MCAR, the absence occurs entirely independently from feature values),
Missing At Random (MAR, the absence depends only on the observed feature values),
and Missing Not At Random (MNAR, the absence depends on both observed and the unobserved feature values)~\cite{GeneratingMissing}.

Within this work, we consider high-dimensional data with categorical features. An example of such a situation is
Single Nucleotide Polymorphism (SNP) data, which presents a range of genetic differences between the individuals. Typically the associations between traits/diseases are studied. 
A standard coding for values in SNP datasets is 0, 1, and 2 for variants $AA$, $Aa$, and $aa$ respectively, where allele $A$ corresponds to the prevalent variant in the population and allele $a$ to the minor one. Due to linkage disequilibrium~\cite{Das2018}, 
neighboring features can correlate to each other, and taking such dependencies into account is helpful for missing value imputation. At the same time, some long-distance correlations (across the genome) are also possible, though rare. A typical SNP dataset contains a number of features (positions on the genome) greatly exceeding the number of samples (individuals in the population of the study). SNP datasets are prone to a missing values problem due to a variety of reasons, such as deviations from the Hardy-Weinberg equilibrium, an abundance of rare variants, and missing features in combining different datasets in meta-studies~\cite{Das2018}.
Within this study, we assume that missing data is missing completely at random as it depends on external factors rather than observed/unobserved values.
Removing features or samples containing missing values may be inefficient as this implies loss of important information and impoverishment of the data.

For SNP data, imputation methods can be split into reference-based and reference-free methods.
Reference-based techniques require a reference panel based on whole-genome sequencing samples and show the advantage of using large datasets with complete data as well as additional information such as linkage patterns, mutations, and recombination hotspots~\cite{Das2018}.
The main limiting factors for such methods are the size and sequencing coverage of reference panels, as well as the conformity of ethnicity in the reference panel and data containing missing values to impute. While reference-based methods are considered a first-choice approach to impute missing values in SNP data, the corresponding reference panels are not always available, especially for non-human species. Moreover, these methods require similarity of populations in the references and data to impute. These facts necessitate the study and development of methods that are independent of the reference panels and impute missing values exploiting only statistical inferences from the data itself.

The existing missing value imputation methods range from simple replacement with mean, mode, or median statistics~\cite{LittleRubin} to more sophisticated techniques such as k-Nearest Neighbours (kNN)~\cite{kNN}, Singular Value Decomposition (SVD)~\cite{SVD}, Random Forests~\cite{MissForest}, and logistic regression. Recently developed deep learning techniques have also been applied for imputation, e.g. Denoising Autoencoders~\cite{SCDA} method. Below we present the listed methods in more detail.

{\bf Mode}~\cite{LittleRubin}.
For each feature, a mode of non-missing values, i.e. the most frequent value, is estimated, and the missing values are imputed with this mode.

{\bf k-Nearest Neighbors (kNN)}~\cite{kNN}.
The imputation procedure is based on the weighted k-Nearest Neighbors algorithm. The algorithm looks for the $k$ samples that are most similar to the one whose missing values need to be replaced and uses these $k$ neighbors to impute the missing values. For experiments, we used \textsf{knncatimpute} function implemented in \textsf{scrime} R package.

{\bf Singular Value Decomposition (SVD)}~\cite{SVD}.
This method calculates the $k$ most significant
eigenvectors and then imputes the missing values using a low-rank SVD approximation estimated by an Expectation-Maximization algorithm.
For experiments, we used \textsf{IterativeSVD} function implemented in \textsf{fancyimpute}~\cite{fancyimpute} python package.

{\bf Multivariate Imputation by Chained Equations (MICE)} ~\cite{MICE}. 
The imputation process is organized into several cycles of prediction, on one cycle each variable is regressed on the other variables (all or subset).
Initially, MICE imputes missing values with samples from features distributions and then carries out a number of imputation iterations. The MICE method is very flexible w.r.t. base model, i.e. any per feature estimator is possible.

{\bf MissForest}~\cite{MissForest}. MissForest also works in an iterative manner by predicting missing values by Random Forests trained on the observed features. 
MissForest builds a new Random Forest for each feature and iteration, while we propose to use a much smaller number of forests with a small number of features each, and this allow to significantly alleviate the computation (see Subsection~\ref{subsec:time})

The MICE and missForest methods are commonly used for different types of data and, in particular, clinical data, and can be considered state-of-the-art for missing value imputation, but we have not found big evidence of using these methods for high-dimensional datasets, as they become very costly with the rise of the number of features. In our empirical study we try to adapt the MICE method for SNP data, but do not obtain promising results (see Section~\ref{sec:results}).

\begin{figure}
    \centering
		\subfloat[Classic Autoencoder 
        \label{fig:AE_simple}]
            {\includegraphics[width=0.45\textwidth]{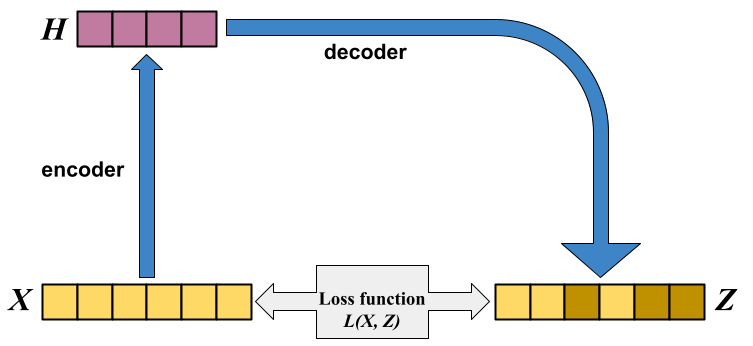}}  \quad 
        \subfloat[Denoising Autoencoder \label{fig:AE_denoising}]
            {\includegraphics[width=0.45\textwidth]{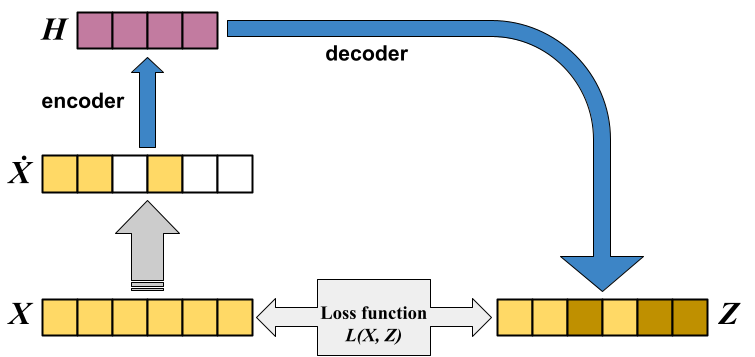}} \\
	\subfloat[Autoencoder without an explicit encoding \label{fig:AE_nonencoding}]
        {\includegraphics[width=0.6\textwidth]{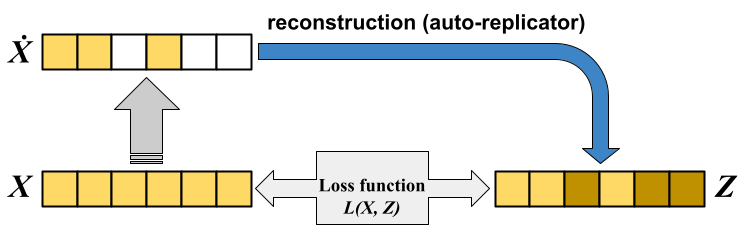}} 
            
	\caption{\footnotesize An illustration of (a) Classic Autoencoder (with hidden representation $H$), (b) Denoising Autoencoder (input is corrupted with noise or missing values as $\dot{X}$ before encoding), and (c) Autoencoder without an explicit encoding (as we use in our work). 
	In all cases, the goal is to minimize the difference between $X$ and its reconstruction $Z$.
    }
\end{figure}

{\bf Denoising Autoencoders (DAE)}~\cite{DAE,SCDA}.
Neural networks reproducing input $X$ as output are called Autoencoders~\cite{AE}. 
Classical Autoencoders implemented within neural networks architecture consist of Encoder and Decoder structures as illustrated in Fig.~\ref{fig:AE_simple}. 
While the inner structure of hidden layers can be very different, the typical common property is having a narrow middle layer $H$ to restrict the model to learning only important information from the data.
Optimizing hidden layers implies a search for some inner patterns in the data.
Denoising Autoencoders~\cite{DAE} receive data $\dot{X}$ corrupted by noise or missing values as input and complete data $X$ as output during the training phase when they learn to remove noise or impute missing values (see Fig.~\ref{fig:AE_denoising}). Denoising Autoencoders have been successfully applied to address the missing data problems in various fields~\cite{DAE}.
In~\cite{SCDA} the authors suggest Sparse Convolutional Denoising Autoencoders (SCDA) to impute missing values in SNP datasets. Sparsity is required due to high dimensionality and insufficient number of samples to train on, and convolution layers are used as neighboring features have a bigger chance to explain each other, at least in SNP data. 
The main limitation of the SCDA method is that they require training data of complete cases, which is usually very limited in SNP data. For this reason, we don't include the SCDA method in an experimental setting where all or almost all cases are affected by missingness.

Although apparently largely overlooked in the literature, we have noticed 
that any other model designed for the multi-label prediction can be used instead of a neural network as an Autoencoder. 
One such example is a combination of decision trees~\cite{treeAE} where the first decision tree is used as an encoder, and the second one is used in a vice versa manner as a decoder.
Meanwhile, this idea can be simplified even more: if we are not aiming to extract the information about the inner patterns of the data, the usage of a straightforward model such as a decision tree or random forest is sufficient.
Such an approach can facilitate the optimization process for the model on data containing a small number of samples, and at the same time, decision trees and random forests are both efficient and simple to understand and interpret.
To the best of our knowledge, this simple but efficient idea has not been well studied in the literature. We argue, that however it deserves attention and can be further investigated. Applying this idea, we suggest further Autoreplicative Random Forests.
While we choose Random Forests as a well-known and stable multi-label method with good performance, this idea may be developed by using other multi-label methods, such as e.g. Classifier Chains~\cite{CCReview}, Multilabel k Nearest Neighbours~\cite{MLkNN}, Random k-Labelsets~\cite{RakEL}, Conditional Dependency Networks~\cite{CDN}.


\section{Method}
\label{sec:method}

Our method consists of two main novelties.
First, 
we use multi-label classifiers (e.g. Random Forests) as imputational Autoreplicative models (Subsection~\ref{subsec:method_ae}).
Second,
Chains of subsequent windows of Autoreplicative models 
allow adapting the idea of Autoencoders to real-world high-dimensional scenarios when there is no complete data available for training (Subsection~\ref{subsec:method_windows}). 


\subsection{Autoencoders without an explicit encoding}
\label{subsec:method_ae}

Our method is inspired by the idea of Denoising Autoencoders for missing value imputation. We use multi-label predictive models that are not based on neural networks. With this approach, we want to efficiently process relatively low-sampled (compared to a number of features) datasets, where complex neural networks are prone to overfitting and get stuck in optimizing parameters. Furthermore, as discovering the inner structure of the data itself is out of the scope of this task, we do not explicitly need hidden layers of the neural network.
We would like to point here that any multi-label classification model can be used for this goal. We will use Random Forest as it proved to be a competitive and robust method. However, an approach of autoreplicative imputation models deserves better research and possibly may be improved by the usage of more sophisticated multi-label models. 
To the best of our knowledge, multi-label classification models and, e.g., Random Forests have not been used before as autoreplicative models reproducing input as output.

The training process is illustrated in Fig.~\ref{fig:AE_nonencoding}. First, we select complete cases of the entire dataset or of a features subset (more on that in the following subsection) $X$, obtain dataset $\dot{X}$ by manually corrupting them with missing values
(uniformly distributed, following the proportion of missing values in the original dataset)
, and train an Autoreplicative forest to reproduce $Z \sim X$ from $\dot{X}$, i.e. fill missing values by minimizing loss function between $Z$ and $X$. Then the fitted model can be used to replace actual missing values.


\subsection{Ensemble of chains of Autoencoders}
\label{subsec:method_windows}

Autoencoders are considered a baseline method for missing value imputation. However, they require complete data for training. 
It is often difficult or impossible to obtain such datasets in real-world problems when missing values can be abundant.
This is especially the case for high-dimensional data: with a large number of features, it is likely not feasible to select a reasonable number of rows without missing values, even for a small ratio of missingness. 
As a consequence, we need to adapt our approach for the high-dimensional datasets, when training data may be not available. 
For this goal, we split the whole set of consequent features into windows of size $\Delta$, such that for the features within one window it is possible to select a training subset of reasonable size with full information. We fit the model on the selected subset and then predict values to impute missing ones in the remaining subset. 

In the case of SNP data, it makes sense to select windows of consequetive features, as they are more likely to provide information about each there. This effect is due to linkage disequilibrium, i.e. close neighbor positions in the genomes are more likely to be inherited from the same ancestors. This window approach may serve for other types of ordered data, such as e.g. gene expression arrays, time series, images, and sound fragments.

\begin{figure}[H]
        \centering
        \includegraphics[width=0.7\textwidth] 
        {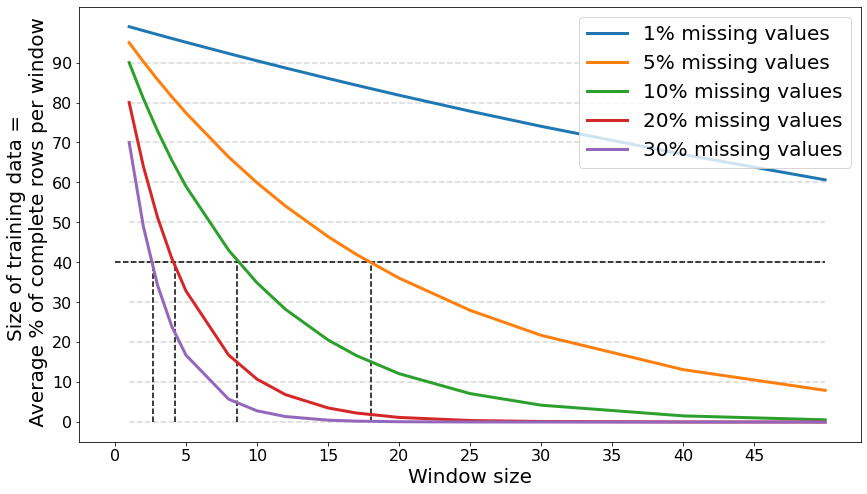}
        \caption{\footnotesize Average complete training size (i.e. rows without missing values) according to window size $\Delta$.
        Missing values are simulated for the MCAR scenario with a uniform distribution.
        Dashed black lines show examples of possible window sizes for $\tau = 0.4$. 
        }
        \label{fig:windowsize}
\end{figure}

\begin{figure}
    \centering
    \includegraphics[width=0.7\textwidth]{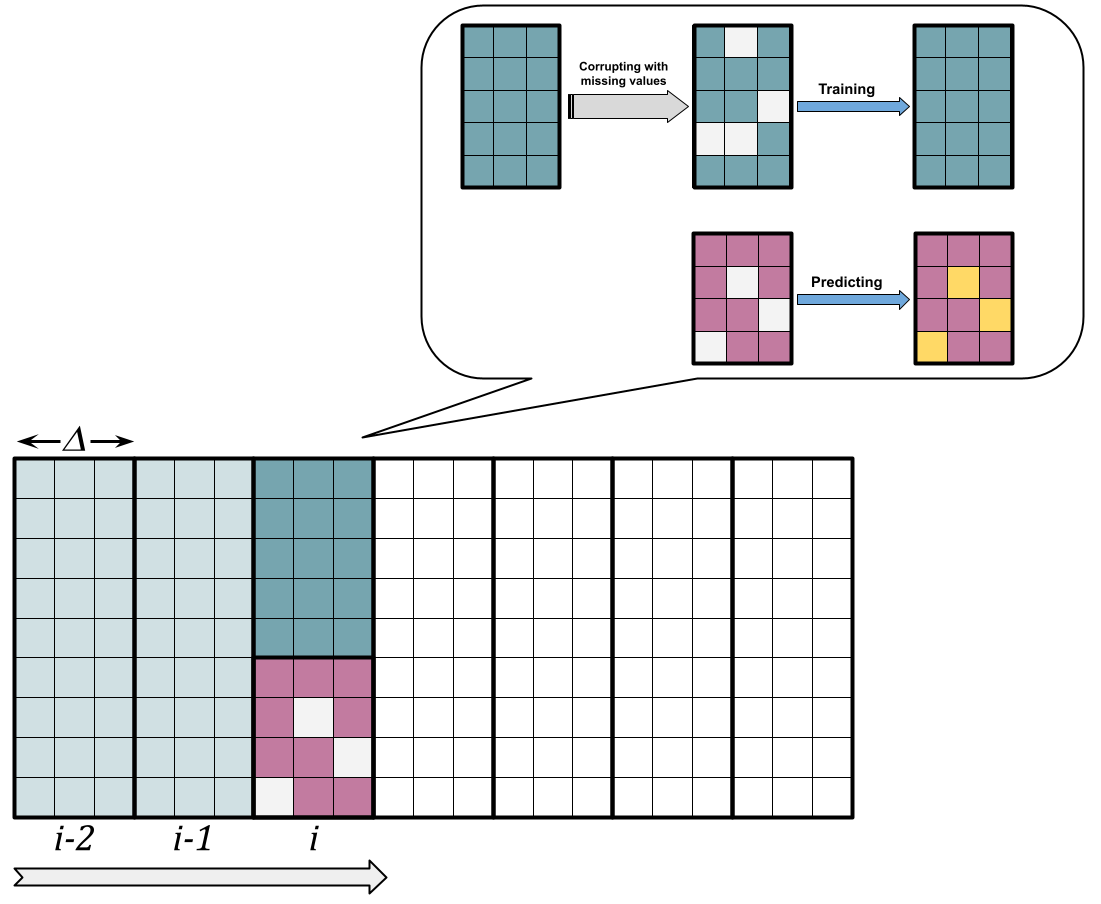}
    \caption{\footnotesize Model processes windows in a chain, incorporating windows with already imputed values as additional features. At one step, we split the window of size $\Delta$ into training part with complete data and testing part with missing values. After fitting on training data corrupted with missing values, we impute missing values in testing part.}
    \label{fig:model}
\end{figure}

Fig.~\ref{fig:windowsize} shows the average size of available training data in simulation with uniformly distributed missing values. As it can be seen, it decreases dramatically with the growth of window size. To increase the method's power to catch and use dependencies between the features, we suggest chains of imputation models, similar to the Classifier Chains methodology~\cite{CCReview}, i.e. stacking of already processed features as new features for the consequent estimators (see Fig.~\ref{fig:model}). 
To keep the complexity of the algorithm feasible and reduce computation time, we do not incorporate all previous windows but select only $\nu$ last ones.

\begin{figure}

    \begin{tabular}{cc}
        \centering
        \subfloat[\footnotesize Probabilities to take already predicted windows into chain at position 50;
        100 windows; 10 previous windows for each chain; 3 chains: forward, backward, and random.
        \label{fig:window_neighbours}]{
        \includegraphics[width=0.5\textwidth]{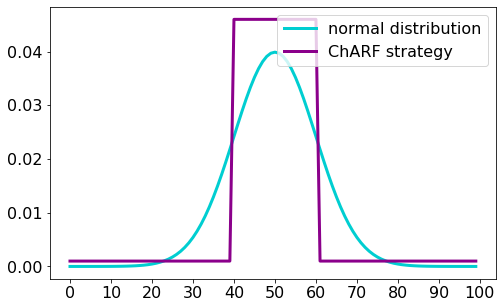}}
        
        \hspace{2pt}

        \subfloat[\footnotesize Two strategies to include distant windows into analysis
        \label{fig:two_strategies}]{
        \includegraphics[width=0.5\textwidth]{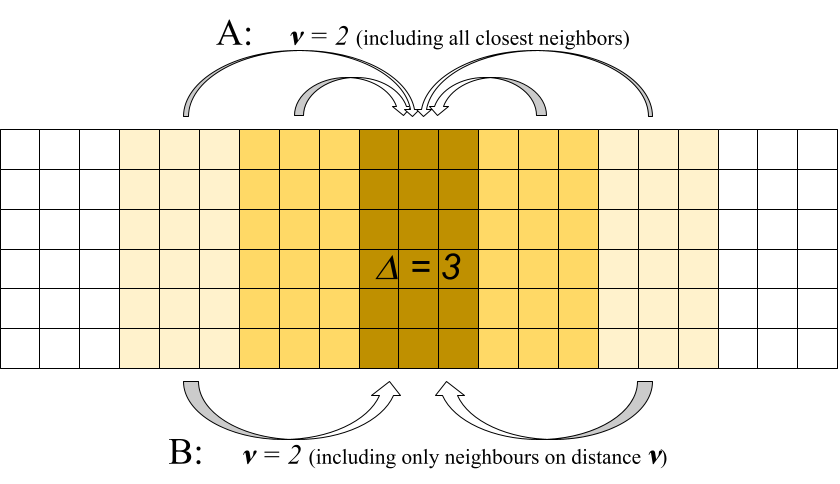}}
    \end{tabular}

    \begin{tabular}{cc}
         \subfloat[\footnotesize Window size $\Delta=10$, strategy $A$
         \label{fig:two_strategies_10_A}]{
            \includegraphics[width=0.5\textwidth] {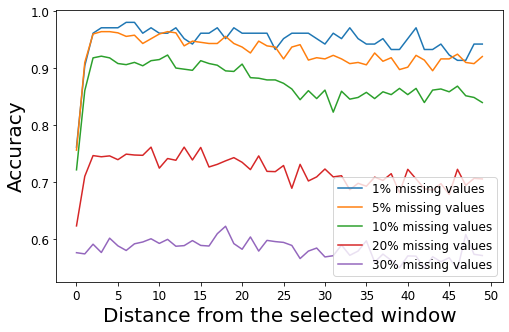}} &
            
         \subfloat[\footnotesize Window size $\Delta=5$, strategy $A$
         \label{fig:two_strategies_5_A}]{
             \includegraphics[width=0.5\textwidth] {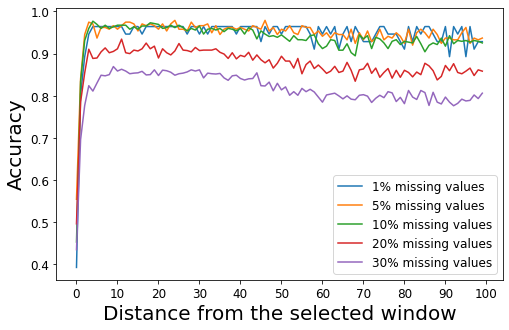}} \\
             
         \subfloat[\footnotesize Window size $\Delta=10$, strategy $B$
         \label{fig:two_strategies_10_B}]{
             \includegraphics[width=0.5\textwidth] {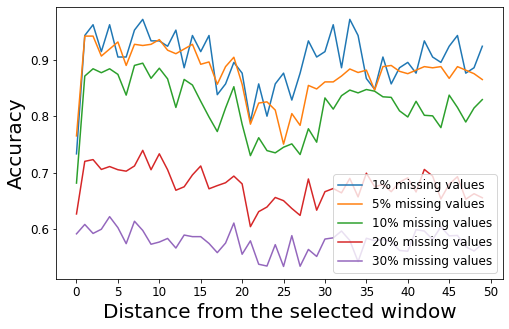}} &
             
         \subfloat[\footnotesize Window size $\Delta=5$, strategy $B$
         \label{fig:two_strategies_5_B}]{
             \includegraphics[width=0.5\textwidth] {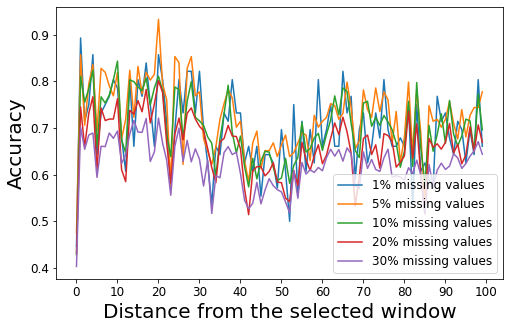}} \\             
    \end{tabular}
    \caption{\footnotesize Including the $2\nu$ closest neighbor windows as additional features (strategy $A$) significantly increases the accuracy while including only 2 windows on distance $\nu$ (strategy $B$) has occasional and unstable improvement.}
    \label{fig:nu_simulation}
\end{figure}

The basic intuition behind using windows of consequent features is that short chromosome segments can be inherited from a distant common ancestor~\cite{Das2018} and thus shared between some individuals.
For this reason, we select one forward and one backward chain, as well as several (up to 3) random chains, to incorporate possible long-term interactions. Selection of previously imputed windows can be generalized as, for example, sampling from a normal distribution with a mean equal to the current window number (Fig.~\ref{fig:window_neighbours}) or other kinds of distributions for different kinds of data.
In the ensemble of chains, we average predictions (i.e. take a major vote) for each missing value.

To support the hypothesis that neighboring features have a higher chance to explain each other, in Fig.~\ref{fig:two_strategies}-\ref{fig:two_strategies_5_B} we include experiments for using all neighboring (strategy A) or only two distant (strategy B) windows on distance $\nu$. We can see that including very close neighbors significantly increases the quality of imputation, while with including distant neighbors the improvement may present (this fact corresponds to possible long-term correlations), but is very unstable and cannot be guaranteed.

\begin{table}[]
    \caption{\footnotesize Example window sizes $\Delta$ according to desired training samples, via Eq.~\ref{eq:delta}}
    \label{tab:Delta}
	\centering
    \begin{tabular}{c|c@{\hskip 8pt}c@{\hskip 8pt}c@{\hskip 8pt}c@{\hskip 8pt}c}
        \toprule
        \backslashbox{Size of training data}{\% of missing data}
        & 1\% & 5\% & 10\% & 20\% & 30\% \\
        \midrule
        20\% of original data & 160 & 31 & 15 & 7 & 4 \\
        30\% of original data & 120 & 23 & 11 & 5 & 3 \\
        50\% of original data & 69 & 14 & 7 & 3 & 2 \\
        \bottomrule
    \end{tabular}
\end{table}

Our method is summarised as pseudocode in Alg.\ref{alg:windowAEimputation}.
We compare performance of the models with hyperparameters $\Delta$ and $\nu$ in Section~\ref{sec:results}.
We estimate theoretically the maximum size $\Delta$ of one window 
according to the desired size of training data $\tau$. As $\tau \sim (1-f)^{\Delta}$, then
\begin{equation}
\label{eq:delta}
\Delta (\tau) \sim \log_{1-f} \tau = \frac{\ln \tau}{\ln(1-f)}, \quad \tau \in (0,1),
\end{equation}
where $f$ is a fraction of missing values and $\tau$ is a desirable threshold for a ratio of complete rows in the training subset. The empirical results of the simulation (Fig.~\ref{fig:windowsize}) correspond to this estimation. We see that with the growth of window size the size of training data decreases dramatically. As a consequence, the window size should be selected carefully by taking the missing value ratio into account. We suggest possible window sizes according to the desired size of training data in Table~\ref{tab:Delta}.

\begin{algorithm}
	\caption{}
    \label{alg:windowAEimputation}
    \begin{algorithmic}[1]
	
	\Procedure{ChARF}{$X_{N \times p}$, window size $\Delta$, \# of previous steps $\nu$, \# of chains $K$}
	    
	    \State Split features into $\Delta$-wide windows
	    \Comment{\parbox[t]{.27\linewidth}{Last window has size $p \Mod \Delta$}}
	    
        \State Generate $K$ permutations of $(1, 2, ..., n = \ceil{\frac{p}{\Delta}}) \}$
        
        \For{each permutation $\{ \sigma(1), ..., \sigma(n) \}$}
            \For{each window  $X_{\sigma(i)}$}
                \State $X_{ext} \gets X_{\sigma(i)} \bigoplus X_{\sigma(i-1)} \bigoplus ... \bigoplus X_{\sigma(i-\nu)}$
                \Comment{\parbox[t]{.27\linewidth}{Stack last $\nu$ processed windows as additional features }}

                \State $X_{train} \gets X_{ext}^{complete}$
                \Comment{\parbox[t]{.27\linewidth}{Select complete cases for training}}
                
                \State $\Tilde{X}_{train} \gets X_{train}$ corrupted with missing values
                \Comment{\parbox[t]{.27\linewidth}{Uniformly distributed, \% of m.v. calculated from $X_{\sigma(i)}$}}

                \State $X_{test} \gets X_{ext}^{missing}$ 
                
                \State Fit model on $(\Tilde{X}_{train}, X_{train})$
                
                \State $X_{pred} \gets $ predictions of fitted model on $X_{test}$
                
                \State replace missing values in $X_{test}$ with corresponding values from $X_{pred}$
            \EndFor
        \EndFor
    \State Take major vote for all $K$ predictions per missing value
    \EndProcedure
\end{algorithmic}
\end{algorithm}


\section{Results and discussion}
\label{sec:results}

\begin{figure}

        \centering
        \includegraphics[width=0.35\textwidth] 
        {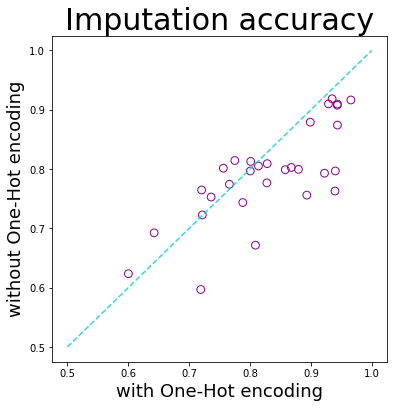}
        \caption{\footnotesize One-Hot encoding may significantly improve the imputation power of ChARF method in SNP datasets. 
        }
        \label{fig:OneHot}

\end{figure}

We evaluate the performance of our method by imputation accuracy, i.e. percentage of correctly imputed values out of missing ones. 
We test Chains of autoreplicative Random Forests (of 10 trees each) on several high-dimensional SNP datasets ($p \gg N$), briefly summarized in Table~\ref{tab:datasets}.
For the SNP data, we test only the MCAR scenario, as missing values are likely to happen because of external reasons rather than depending on missing or observed data.
We simulate the missing values by masking true values in the data under a uniform distribution, with the proportion of missing values 1\%, 5\%, 10\%, 20\%, and 30\%. The procedure is repeated 5 times to produce independent incomplete datasets. Average accuracy is shown.
The empirical study has shown a significant improvement, when the features are one-hot encoded (paired t-test statistics 3.442, df=29, p-value=0.0018, see Fig.~\ref{fig:OneHot}).

\begin{table}
    \caption{\footnotesize Datasets used in experiments, $p$ features, $N$ samples.}
    \label{tab:datasets}
    \centering
    \begin{tabular}{l@{\hskip 8pt}c@{\hskip 8pt}c@{\hskip 8pt}c@{\hskip 8pt}c}
        
        \toprule
         Name & $p$ & $N$  \\
         
         \midrule
         
         Maize~\cite{MaizeData} & 44,729 & 247 \\
         Eucalyptus~\cite{EucalyptusData} & 33,398 & 970 \\
         Colorado Beetle~\cite{ColoradoData} & 34,186 & 188 \\
         Arabica Coffee~\cite{CoffeeData} & 4,666 & 596 \\
         Wheat (Zuchtwert study)~\cite{ZuchtwertData} & 9,763 & 388 \\%
         Coffea Canephora~\cite{CoffeaCanephoraData} & 45,748 & 119 \\
         \bottomrule
    \end{tabular}
\end{table}

\begin{figure}
        \centering
        \begin{tabular}{cccccc}
            & 1\% & 5\% & 10\% & 20\% & 30\% \\   
            
            \makecell{\rotatebox{60}{\tiny Maize}}
                & 
            \makecell{
            \includegraphics[width=0.18\textwidth] 
                {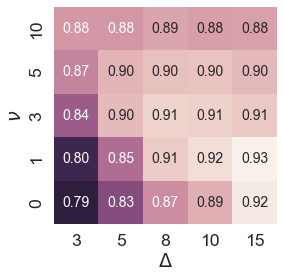}
            }
                & 
                \makecell{
            \includegraphics[width=0.18\textwidth] 
                {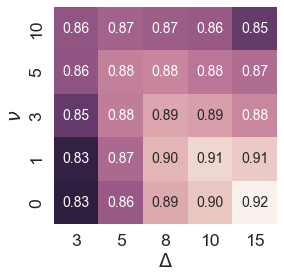}}
                & 
                \makecell{
            \includegraphics[width=0.18\textwidth] 
                {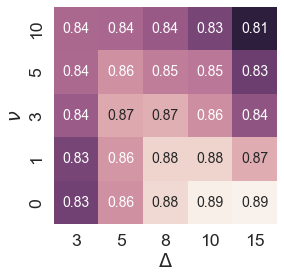}}
                & 
                \makecell{
            \includegraphics[width=0.18\textwidth] 
                {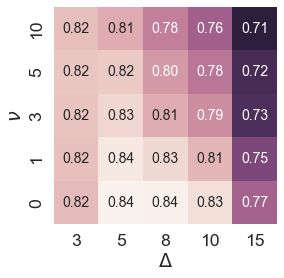}}
                &
                \makecell{
            \includegraphics[width=0.18\textwidth] 
                {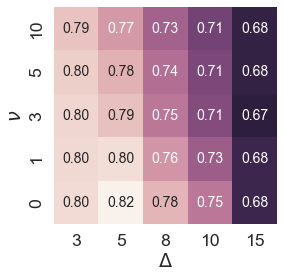} }
                 \tabularnewline
            
            \makecell{\rotatebox{60}{\tiny Eucalyptus}}
                & 
            \makecell{
            \includegraphics[width=0.18\textwidth] 
                {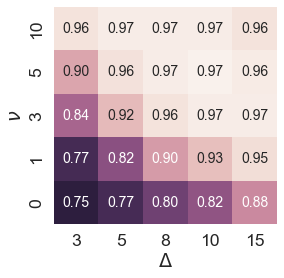}} 
                & 
            \makecell{
            \includegraphics[width=0.18\textwidth] 
                {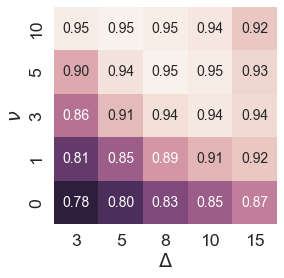}} 
                & 
            \makecell{
            \includegraphics[width=0.18\textwidth] 
                {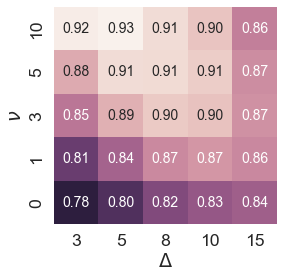}} 
                & 
            \makecell{
            \includegraphics[width=0.18\textwidth] 
                {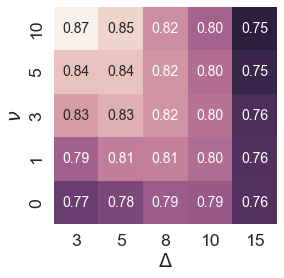}} 
                &
            \makecell{
            \includegraphics[width=0.18\textwidth] 
                {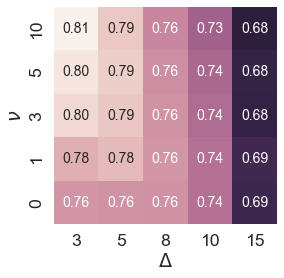}} 
                \\    

            \makecell{\rotatebox{60}{\tiny C. Beetle}}
                & 
            \makecell{
            \includegraphics[width=0.18\textwidth] 
                {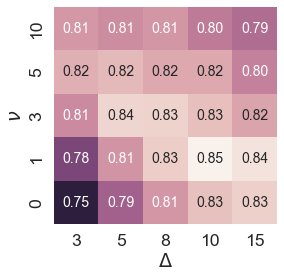}} 
                & 
            \makecell{
            \includegraphics[width=0.18\textwidth] 
                {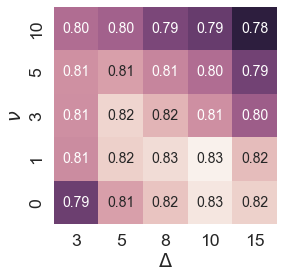}} 
                & 
            \makecell{
            \includegraphics[width=0.18\textwidth] 
                {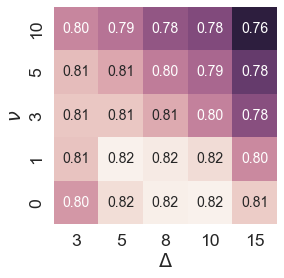}} 
                & 
            \makecell{
            \includegraphics[width=0.18\textwidth] 
                {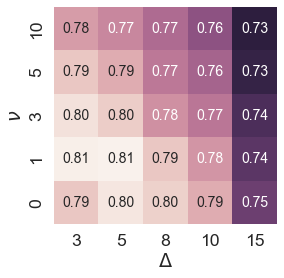}} 
                &
            \makecell{
            \includegraphics[width=0.18\textwidth] 
                {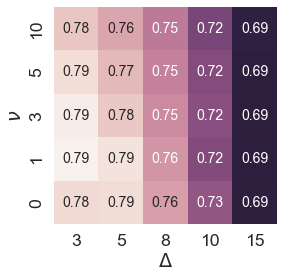}} 
                \\
            
            \makecell{\rotatebox{60}{\tiny A. Coffee}}
                &
            \makecell{
            \includegraphics[width=0.18\textwidth] 
                {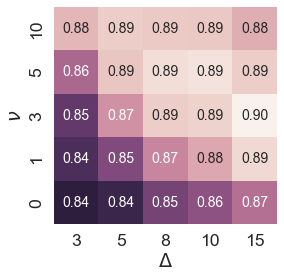}} 
                & 
            \makecell{
            \includegraphics[width=0.18\textwidth] 
                {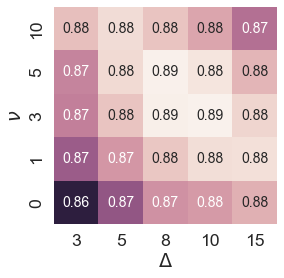}} 
                & 
            \makecell{
            \includegraphics[width=0.18\textwidth] 
                {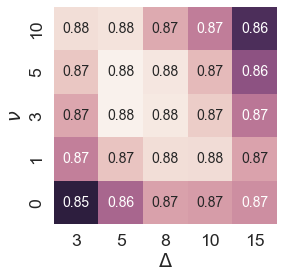}} 
                & 
            \makecell{
            \includegraphics[width=0.18\textwidth] 
                {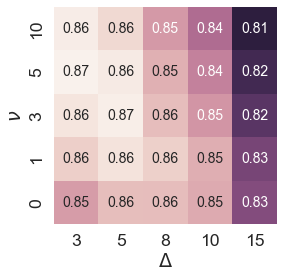}} 
                &
            \makecell{
            \includegraphics[width=0.18\textwidth] 
                {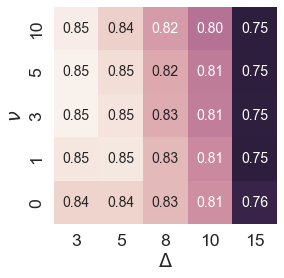}} 
                \\
            
            \makecell{\rotatebox{60}{\tiny Wheat}}
                & 
            \makecell{
            \includegraphics[width=0.18\textwidth] 
                {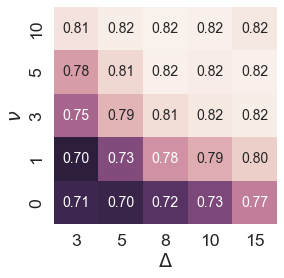}} 
                & 
            \makecell{
            \includegraphics[width=0.18\textwidth] 
                {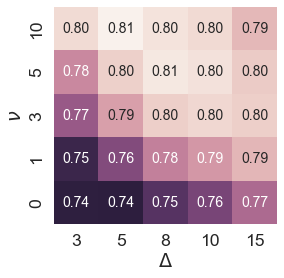}} 
                & 
            \makecell{
            \includegraphics[width=0.18\textwidth] 
                {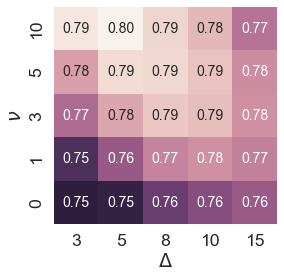}} 
                & 
            \makecell{
            \includegraphics[width=0.18\textwidth] 
                {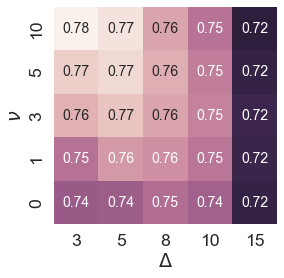}} 
                &
            \makecell{
            \includegraphics[width=0.18\textwidth] 
                {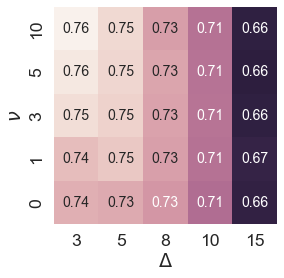}} 
                \\

            \makecell{\rotatebox{60}{\tiny C. Canephora}}
                & 
            \makecell{
            \includegraphics[width=0.18\textwidth] 
                {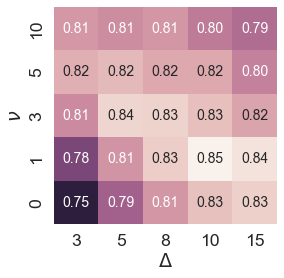}} 
                & 
            \makecell{
            \includegraphics[width=0.18\textwidth] 
                {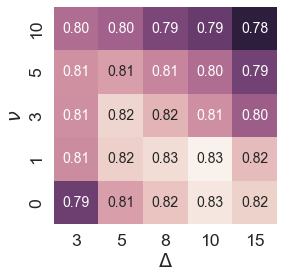}} 
                & 
            \makecell{
            \includegraphics[width=0.18\textwidth] 
                {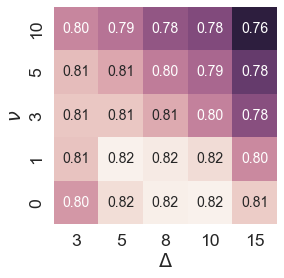}} 
                & 
            \makecell{
            \includegraphics[width=0.18\textwidth] 
                {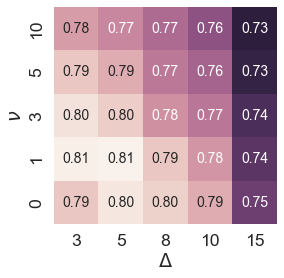}} 
                &
            \makecell{
            \includegraphics[width=0.18\textwidth] 
                {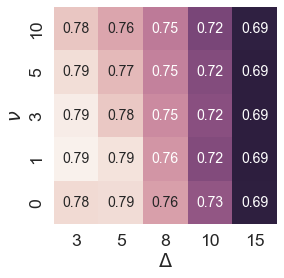}} 
                \\    
        \end{tabular}
        
\caption{\footnotesize Accuracy of imputation for SNP datasets 
for different ratios of missing values (indicated in column headers). Better accuracy lighter/higher value (shown in cells). 
}
\label{fig:gridsearch}
\end{figure}

For ChARF, we first evaluate hyperparameters: 
window size $\Delta \in [3,5,8,10,15]$, and
number of previous windows in the chain to take as new features $\nu \in [0,1,3,5,10]$. 
To reduce the computation time, 
we first search for the best values of $\Delta$ and $\nu$ on reduced datasets (first 1000 features) and then use these values for computation on the entire datasets.
The grid-search results are presented in Fig.~\ref{fig:gridsearch}. As expected (from estimation in Subsection~\ref{subsec:method_windows}), from Fig.~\ref{fig:gridsearch} we see that the most effective window size decreases with the growth of a number of missing values (since a bigger part of instances gets corrupted). 

For the MICE method, with the default settings, each estimator considers all other variables, which makes the total complexity at least quadratic and thus requires huge computational and time resources in the case of high-dimensional data (in our experiments, the machine ran out of memory). The original R package suggests a pre-processing \textsf{quickpred} function, which selects the predictive features based on pairwise correlation, but in this case, quadratic complexity is required in this step. With the intuition that the neighboring features in SNP data have the highest chance to explain each other, for the experiments we select windows of 10 neighbors for each feature. Such an approach is computationally feasible, but the imputation still leaves some missing values in the data (around 10-20\%). The possible explanation is the collinearity between features~\cite{MICE}. This approach worked for smaller SNP datasets (Arabica Coffee and Wheat), but for the other ones, the computations still failed.

\begin{table}[H]
    \tiny
	\caption{\footnotesize Accuracy. An asterisk ($^*$) indicates optimal hyperparameters estimated via internal validation. For knnimpute we selected the best of $k \in \{3, 5, 10, 20, 50\}$ (shown in brackets) in a similar way.   
	Likewise for rank $\in \{10, 20, 50, 100, 200, 300, 500\}$ for SVD method. 
	The missForest method was run for first 100 features only as it is not possible to run it for a whole dataset.
	Best accuracy per column in \textbf{bold}.
    All results rounded to 3 dp. 
    }
    \label{tab:Accuracy}
    \centering
    \begin{tabular}{c|ccccc|ccccc}
        \toprule
            & 0.01 & 0.05 & 0.1 & 0.2 & 0.3 & 
              0.01 & 0.05 & 0.1 & 0.2 & 0.3 \\
        \bottomrule
        
        
        \toprule
         & \multicolumn{5}{c}{Maize} & \multicolumn{5}{c}{Eucalyptus} \\
        \midrule
           & $\Delta=15^*$ & $\Delta=15^*$ & $\Delta=10^*$ & $\Delta=5^*$ & $\Delta=5^*$ &
           $\Delta=10^*$ & $\Delta=5^*$ & $\Delta=5^*$ & $\Delta=3^*$ & $\Delta=3^*$ \\
           
         ChARF  & $\nu=1^*$ & $\nu=1^*$ & $\nu=1^*$ & $\nu=1^*$ & $\nu=1^*$ &
          $\nu=5^*$ & $\nu=10^*$ & $\nu=10^*$ & $\nu=10^*$ & $\nu=10^*$ \\ 
          
          & {\bf 0.952} & {\bf 0.935} & {\bf 0.916} & {\bf 0.882} & {\bf 0.845} & {\bf 0.970} & {\bf 0.950} & {\bf 0.926} & {\bf 0.866} & 0.810 \\
          
         kNN (5/10) & 0.803 & 0.802 & 0.801 & 0.798 & 0.794 & 0.851 & 0.849 & 0.847 & 0.843 &  {\bf 0.839} \\
         
         mode & 0.727 & 0.727 & 0.726 & 0.727 & 0.726 & 0.725 & 0.732 & 0.731 & 0.730 & 0.729 \\
         
         SVD (50/500) & 0.647 & 0.648 & 0.645 & 0.643 & 0.636 & 0.788 & 0.788 & 0.788 & 0.785 & 0.780 \\ 
         
         MICE & -- & -- & -- & -- & -- &
         -- & -- & -- & -- & -- \\
         
         missForest & 0.662 & 0.650 & 0.622 & 0.593 & 0.580 &
         0.684 & 0.673 & 0.626 & 0.564 & 0.521 \\
         \bottomrule
        \noalign{\vskip 2mm}

        
         \toprule
         & \multicolumn{5}{c}{Colorado Beetle} & \multicolumn{5}{c}{Arabica Coffee} \\
         \midrule
           & $\Delta=10^*$ & $\Delta=10^*$ & $\Delta=5^*$ & $\Delta=5^*$ & $\Delta=3^*$ & $\Delta=15^*$ & $\Delta=10^*$ & $\Delta=5^*$ & $\Delta=3^*$ & $\Delta=3^*$ \\
           
         ChARF & $\nu=1^*$ & $\nu=1^*$ & $\nu=1^*$ & $\nu=1^*$ & $\nu=1^*$ & $\nu=3^*$ & $\nu=3^*$ & $\nu=5^*$ & $\nu=10^*$ & $\nu=3^*$ \\
         
          & {\bf 0.835} & {\bf 0.824} & {\bf 0.818} & {\bf 0.805} & {\bf 0.792} & {\bf 0.897} & {\bf 0.886} & {\bf 0.878} & {\bf 0.866} & 0.854 \\
          
         kNN (50/10) & 0.765 & 0.763 & 0.765 & 0.765 & 0.764 & 0.867 & 0.866 & 0.866 & 0.865 & {\bf 0.864}  \\
         
         mode & 0.761 & 0.760 & 0.762 & 0.761 & 0.761 & 0.807 & 0.804 & 0.805 & 0.805 & 0.804 \\
         
         SVD (50/100) & 0.740 & 0.737 & 0.737 & 0.735 & 0.734 & 0.693 & 0.694 & 0.696 & 0.692 & 0.690 \\
         
         MICE & -- & -- & -- & -- & -- &
         0.757 & 0.741 & 0.724 & 0.689 & 0.664 \\
         
         missForest & 0.352 & 0.349 & 0.361 & 0.326 & 0.335 &
         0.497 & 0.480 & 0.533 & 0.541 & 0.586 \\
         
         \bottomrule
        \noalign{\vskip 2mm}  
        

         \toprule
         & \multicolumn{5}{c}{Wheat} & \multicolumn{5}{c}{Coffea Canephora} \\
         \midrule
           & $\Delta=8^*$ & $\Delta=5^*$ & $\Delta=5^*$ & $\Delta=3^*$ & $\Delta=3^*$ & $\Delta=10^*$ & $\Delta=10^*$ & $\Delta=5^*$ & $\Delta=5^*$ & $\Delta=3^*$ \\
           
         ChARF  & $\nu=10^*$ & $\nu=10^*$ & $\nu=10^*$ & $\nu=10^*$ & $\nu=10^*$ & $\nu=1^*$ & $\nu=1^*$ & $\nu=1^*$ & $\nu=1^*$ & $\nu=1^*$ \\ 
         
          & 0.821 & 0.808 & 0.795 & 0.777 & 0.762 & {\bf 0.799} & {\bf 0.781} & {\bf 0.761} & 0.731 & 0.717 \\
          
         kNN (10/10) & {\bf 0.823} & {\bf 0.819} & {\bf 0.818} & {\bf 0.815} & {\bf 0.811} & 0.737 & 0.739 & 0.737 & {\bf 0.734} & {\bf 0.731} \\
         
         mode & 0.729 & 0.727 & 0.729 & 0.729 & 0.727 & 0.691 & 0.693 & 0.692 & 0.692 & 0.691 \\
         
         SVD (200/50) & 0.622 & 0.618 & 0.609 & 0.600 & 0.594 & 0.456 & 0.453 & 0.450 & 0.449 & 0.450 \\
         
         MICE & 0.641 & 0.635 & 0.621 & 0.585 & 0.545 & 
         -- & -- & -- & -- & -- \\
         
         missForest & 0.614 & 0.736 & 0.746 & 0.756 & 0.755 &
         0.377 & 0.449 & 0.442 & 0.395 & 0.383 \\
         
         \bottomrule
        \noalign{\vskip 2mm}


    \end{tabular}
\end{table}

In most cases, the experiments show better or competitive performance w.r.t. benchmark methods (Table~\ref{tab:Accuracy}). At the same time, we see that with the rise of the missing value ratio the accuracy of imputation diminishes. This is explained by the very small size of training data even on small windows for a big number of missing values. However, for a moderate missing value ratio, our method consistently outperforms its alternatives.


\subsection{Time complexity analysis}
\label{subsec:time}


The complexity of ChARF is $\mathcal{O} (p N \log N)$ w.r.t. the number of features for a single tree and for an ensemble is $\mathcal{O} (\frac{p}{\Delta} \cdot \Delta N \log N) \sim \mathcal{O} (p N \log N)$ (fixed number of chains and ensemble members).

\begin{figure}
        \centering
        \includegraphics[width=0.35\textwidth]{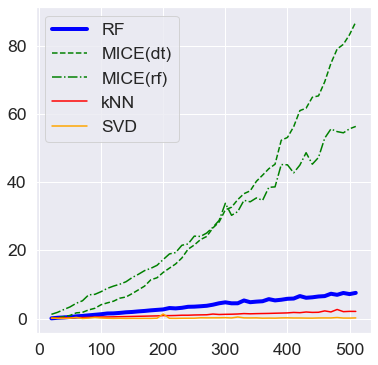}
    	\caption{\footnotesize Empirical results on time complexity for imputation methods.
        }
        \label{fig:Times}
\end{figure}

We expect that the kNN and SVD methods' time complexity is linear w.r.t. a number of features.
For the MICE method time, the complexity depends on the base per feature estimator. In the simulation, we use a decision tree and random forest as base estimators. For a single decision tree, the complexity is $\mathcal{O} (p N \log N)$, and thus total complexity is quadratic w.r.t. the number of features. 
Random forests consist of individual decision trees, but it is possible to select a number of features per tree. The standard choice is $\sqrt{p}$ features per tree, which makes the total complexity $\mathcal{O} (p\sqrt{p})$ and is already substantially slower than linear complexity for a big number of features $p$.
The same estimation holds for the MissForest method.
However, we can reach linear complexity if we put a number of features per tree equal to come constant, which we try in the previous subsection.
These theoretic estimations are well supported in the simulation study, see Fig.~\ref{fig:Times}. We empirically compare the time complexity of the imputation methods on subsets of the Eucalyptus dataset under the MCAR scenario with 10\% missing values. The subsets are selected as first $p_s$ features of the original dataset, $10 < p_s < 500$.

\section{Conclusions and future work}
\label{sec:conclusions}

We propose a new approach: tackling missing value imputation as a multi-label predictive problem. First, we suggest Autoreplicative Random Forests as a simpler alternative for Denoising Autoencoders.
With this simple but efficient idea, we can obtain competitive results, which may be further improved by more sophisticated multi-label models. 
This method does require complete cases for the training phase, though we suppose that in real-world low-dimensional data this scenario is still quite realistic. Besides, it is possible to tune this method by splitting the data into blocks of complete and missing features.

For high-dimensional datasets, when having complete data in all features is very unlikely, we propose Chains of Autoreplicative Random Forests. This method splits data into windows of consequent features and imputes missing values window by window while incorporating information from already processed features. 
We test this approach on SNP datasets and
demonstrate a very competitive predictive power. Our method requires neither reference panels nor complete data for training and thus can be used in a variety of real-world scenarios when imputation of missing data is required. Our approach consists of two main novelties: it is model agnostic (we used using Random Forests in experiments) in regards to the Autoreplicator; essentially an Autoencoder with no explicit encoding; and operates on windows of data. 
Our approach proved competitive, 
and is promising for further investigation. 

In future work, we are going to improve algorithm performance on datasets with a big number of missing values and make it more stable w.r.t.\ high missing value ratios. 
As preliminary experiments show that this approach works for the MAR scenario as well, we will further analyze the performance of the ChARF method for other patterns of missingness. 

We will look at allowing multiple hypotheses and their distribution per a single missing value. This would allow a greater chance of capturing the true mode and incorporating this uncertainty to further data analysis.


\section*{Acknowledgments}

We would like to thank Ander Carre\~{n}o, University of the Basque Country, for fruitful discussions and three anonymous reviewers for their editorial comments.


\bibliographystyle{splncs04}
\bibliography{bib}

\end{document}